\documentclass{article} 
\usepackage{graphicx}
\usepackage[final]{emc2_2020}

\usepackage[utf8]{inputenc}
\usepackage[T1]{fontenc}
\usepackage{hyperref}
\usepackage{url}
\usepackage{booktabs}       
\usepackage{amsfonts}       
\usepackage{nicefrac}       
\usepackage{microtype}  

\begin{document}
\title{FactorizeNet: Progressive Depth Factorization for Efficient Network Architecture Exploration Under Quantization Constraints} 

\author{Stone Yun \\
\texttt{s22yun@uwaterloo.ca} \\
University of Waterloo\\Vision and Image Processing Lab
\And Alexander Wong \\
\texttt{a28wong@uwaterloo.ca} \\
University of Waterloo\\Waterloo Artificial Intelligence Institute
}

\maketitle

\vspace{-0.2in}
\begin{abstract}
\vspace{-0.1in}
Depth factorization and quantization have emerged as two of the principal strategies for designing efficient deep convolutional neural network (CNN) architectures tailored for low-power inference on the edge. However, there is still little detailed understanding of how different depth factorization choices affect the final, trained distributions of each layer in a CNN,  particularly in the situation of quantized weights and activations.  In this study, we introduce a \textit{progressive depth factorization} strategy for efficient CNN architecture exploration under quantization constraints.  By algorithmically increasing the granularity of depth factorization in a progressive manner, the proposed strategy enables a fine-grained, low-level analysis of layer-wise distributions. Thus enabling the gain of in-depth, layer-level insights on efficiency-accuracy tradeoffs under fixed-precision quantization. Such a progressive depth factorization strategy also enables efficient identification of the optimal depth-factorized macroarchitecture design (which we will refer to here as \textbf{FactorizeNet}) based on the desired efficiency-accuracy requirements.
\vspace{-0.05in}
\end{abstract}
\vspace{-0.15in}
\section{Introduction}
\vspace{-0.1in}
\label{sec:intro}
Following the recent explosion in deep learning research, there has been increased attention on complexity reduction strategies for deep convolutional neural networks (CNN) to enable inference on mobile processors. Quantization \cite{DBLP:journals/corr/abs-1712-05877, DBLP:journals/corr/abs-1903-08066, DBLP:journals/corr/abs-1906-04721}, and depth factorization \cite{DBLP:journals/corr/HowardZCKWWAA17, DBLP:journals/corr/ZhangZLS17, DBLP:journals/corr/XieGDTH16, Chol17Xception} have quickly emerged as two highly effective strategies for reducing the power and computational budget needed for on-device inference. These two methods work orthogonally. Fixed point quantization enables simple, low bit-width integer operations which are several times faster/less power than floating point (fp32) operations. Depth factorization reduces the number of CNN parameters and multiply-accumulate (MAC) operations. For depth factorization, we split the input channels into \textit{f} groups and apply \textit{f} groups of filters independently to their respective channel groups. For a given factorization rate, \textit{f}, the number of MACs in a convolution layer goes from (\ref{eq:regularconvmacs}) to (\ref{eq:factorizeconvmacs}), thus reducing computation by a factor of \textit{f}. For simplicity, our equations have excluded the MAC contribution from the pointwise convolution that typically follows the group convolution. Pointwise convolution is often used for the dual purpose of mixing channel information and increasing channel depth.
\vspace{-0.07in}
\begin{equation}
\vspace{-0.1in}
\label{eq:regularconvmacs}
\ K \times K \times H \times W \times C_{in} \times C_{out}\
\end{equation}
\vspace{-0.05in}
\begin{equation}
\label{eq:factorizeconvmacs}
\ K \times K \times H \times W \times \frac{C_{in}}{f} \times \frac{C_{out}}{f} \times f\
\vspace{-0.03in}
\end{equation}
\vspace{-0.12in}

Depthwise separable convolution as described in MobileNets~\cite{DBLP:journals/corr/HowardZCKWWAA17} has become a staple in efficient network design. It represents the extreme end of the depth factorization spectrum with one convolution filter per input channel. However, perhaps we do not always need to go to the extreme. A key tradeoff when designing CNNs for limited compute is efficiency vs. accuracy. As we scale down our architectures, we will necessarily lose accuracy. While depthwise separable convolutions are extremely efficient, they suffer from low data parallelism making them less suited to hardware acceleration. Also as mentioned in \cite{Chol17Xception}, they should not be assumed as the optimal point on the depth-factorization-spectrum. Furthermore, with quantization emerging as essential for on-device inference, we must consider the additional component of quantization error. In general, efficient architectures have so few parameters that they often suffer more quantized accuracy loss compared to higher complexity networks. However, there is still limited understanding of how different architectural choices impact quantized accuracy. Given the significant investment involved with architecture search/design, it would be beneficial to gain detailed insights on the potential quantizability of an architecture during the design phase. Thus, helping speed-up the quantization optimization process. 

\begin{figure*}[t]
\vspace{-0.13in}
\centerline{\includegraphics[width=9cm]{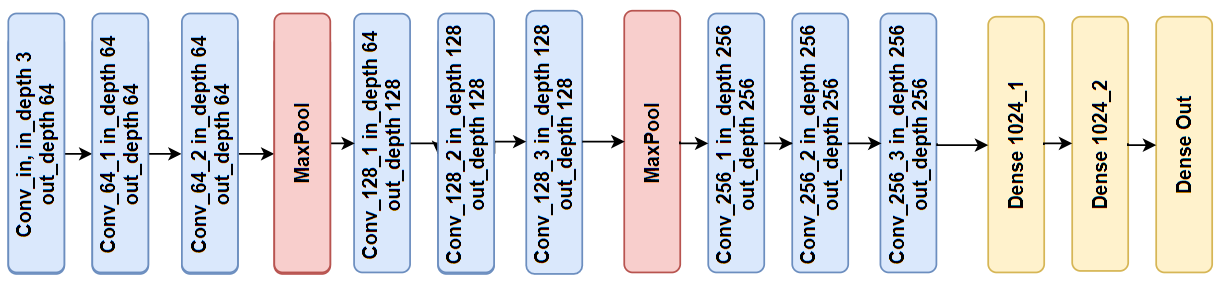}}
\caption{\footnotesize{}\textbf{FactorizeNet Macroarchitecture}. For our progressive, fine-grained analysis we start with a simple regular CNN and fix the macroarchitecture. We then progressively increase the level of factorization of each block using Groupwise Separable Convolution with varying \textit{f}. The very first convolution layer stays fixed.}

\vspace{-0.15in}
\label{fig:factorizenet}
\end{figure*}

We introduce a systematic, progressive depth factorization strategy for exploring the efficiency/accuracy trade-offs of scaling down CNN architectures under quantization and computation constraints. Starting with a simple, fixed macroarchitecture (see Figure~\ref{fig:factorizenet}) we algorithmically increase the granularity of depth factorization in a progressive manner while analyzing the final trained layerwise distributions of weights and activations at each step. Our proposed strategy enables a fine-grained, low-level analysis of layer-wise distributions to gain in-depth, layer-level insights on efficiency-accuracy tradeoffs under fixed-precision quantization. Furthermore, we can identify optimal depth-factorized macroarchitectures which we will refer to as \textbf{FactorizeNet}. While previous studies \cite{DBLP:journals/corr/abs-1711-09224, DBLP:journals/corr/XieGDTH16} have performed ablation studies on the effect of different factorization choices on testing accuracy, they used a high-level approach and were mainly concerned with fp32 accuracy. \cite{8524017} performs layerwise analysis of the signal-to-quantization-noise-ratio (SQNR) to identify layers that were hurting the quantized accuracy of MobileNetsv1 before retraining a modified MobileNets architecture. Our method can be seen as expanding on this approach and going to an even lower level, directly analyzing the distributions at each layer. Insights gained from such a fine-grained approach can help guide further exploration for quantization-based optimizations or provide a baseline expectation of quantized accuracy trade-offs when engineers deploy their quantized model as-is.

\vspace{-0.02in}

\begin{figure}[t]
\vspace{-0.2in}
\centerline{\includegraphics[width=9cm]{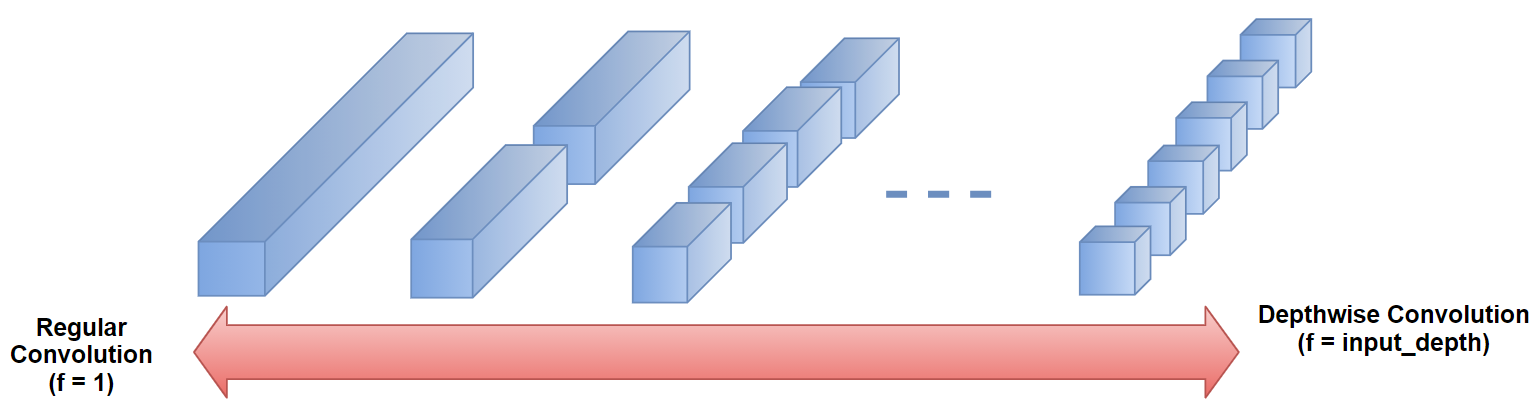}}
\caption{\footnotesize{} The depthwise factorization spectrum. On one end, we have regular convolution, with factorization rate of \textit{f = 1}. On the other end we have depthwise convolution with factorization rate of \textit{f = input-depth}. For a given layer in a CNN architecture, the optimal level of factorization could lie anywhere on this spectrum.}
  \label{fig:depthfactorizespectrum}
  \vspace{-0.01in}
\end{figure}

\vspace{-0.05in}
\section{Progressive Depth Factorization and Fine-Grained Layer Analysis}
\vspace{-0.1in}
\label{sec:progressivedepthfactorize}
Consider a spectrum of depth factorization (see Figure~\ref{fig:depthfactorizespectrum}) with regular convolution on one end (factorization rate \textit{f = 1}) and depthwise convolution on the other (factorization rate \textit{f = input depth}). As we turn the knob from \textit{f = 1} to \textit{f = input depth} for each layer or set of layers in a given macroarchitecture, we will observe a range of efficiency/accuracy trade-offs. Thus, a given CNN macroarchitecture is a search space in itself where a large range of factorization levels and combinations of factorizations can be realized to meet given efficiency-accuracy constraints. Besides searching for the optimal factorization configuration, we also wish to gain detailed insight on the impact of various factorization choices on the layer-wise distributions of final trained weights and activations. This information can help us understand which factorization settings are the most amenable to quantization as well as provide detailed insight on the response of various stages of a CNN to depth factorization. We propose algorithmically increasing the factorization of a given CNN macroarchitecture in a progressive manner while conducting a low-level analysis of the layerwise distributions for each level of factorization. At each factorization step, we train the factorized CNN and track the dynamic ranges of each layer's weights and activations as well as their ``average channel precision". Average channel precision is defined as (\ref{eq:averageprecision}). Channel precision in this context is the ratio between an individual channel's range and the range of the entire layer. \cite{DBLP:journals/corr/abs-1906-04721} algorithmically maximizes the channel precisions of each layer in a network prior to quantization. It can be seen as a measure of how well the overall layer-wise quantization encodings represent the information in each channel. For dynamic ranges of activations, we randomly sample N training samples and observe the corresponding activation responses. To reduce outlier noise, we perform percentile clipping (Eg. top and bottom 1\%) and track the dynamic range and average precision of the clipped activations. As percentile clipping has become a ubiquitous default quantization setting we feel that this method establishes a realistic baseline of what can be expected at inference-time. Finally, there is one more set of dynamic ranges to observe. Batch Normalization (BatchNorm) \cite{DBLP:journals/corr/IoffeS15} has become the best-practice in CNNs. However, their vanilla form is not well-suited for mobile hardware processing. Best practices for mobile inference usually involve folding the scale and variance parameters of BatchNorm into the preceding layer's convolution parameters as described in ~\cite{DBLP:journals/corr/abs-1712-05877}. Therefore, we track the dynamic range and precision of the CNN's batchnorm-folded (BN-Fold) weights.
\vspace{-0.05in}
\begin{equation}
\vspace{0.1in}
\label{eq:averageprecision}
average\_precision = \frac{1}{K}\sum_{i=1}^K \frac{range_{channel\_i}}{range_{tensor}}\
\vspace{-0.2in}
\end{equation}

In this manner, we can iterate through progressively increasing factorization configurations, gaining insights on the efficiency/accuracy trade-offs at each step as well as the final layerwise distributions. Besides enabling analysis of depth factorization, this fine-grained approach is applicable to helping us understand the impact of other architecture choices such as skip/residual connections as well as training hyperparameters such as weight initializations, learning rate schedules etc. Progressive Depth Factorization provides a general framework not only for systematically understanding the efficiency/accuracy trade-offs of factorization, but also for finding the optimal factorization configuration. As there are many directions that can be taken through the ``Progressive Depth Factorization space", our method can be merged with automated search methods such as GenSynth \cite{8822909} to trace out various paths through the space, especially for increasingly complex architectures.

\begin{figure*}
\vspace{-0.2in}
\centerline{\includegraphics[width=9cm]{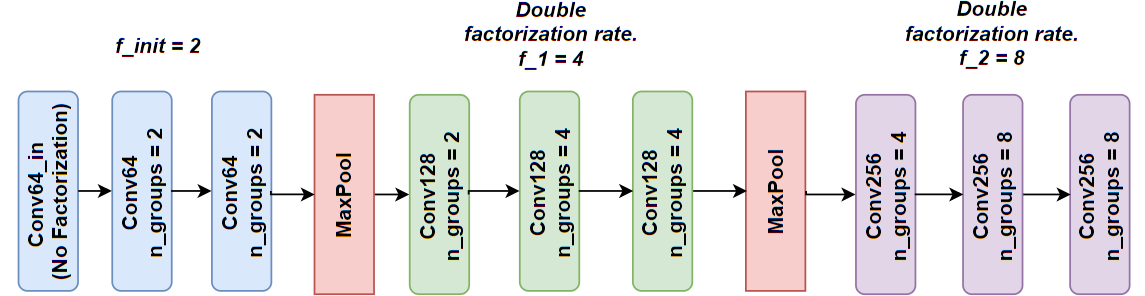}}
\vspace{-0.1in}
\caption{\footnotesize{} \textbf{Reverse Pyramid Factorization Scheme}. For this factorization scheme, we start with an initial factorization rate, f\textsubscript{init}, and double the factorization rate each time the input depth doubles, thus preserving the number of channels per group throughout the network. For f\textsubscript{init} = 64, we recover the depthwise separable CNN.}
\vspace{-0.13in}
  \label{fig:reversepyramid}
\end{figure*}

\begin{figure*}
    \centering
    \begin{minipage}{0.25\textwidth}
        \centering
        \includegraphics[width=0.99\textwidth]{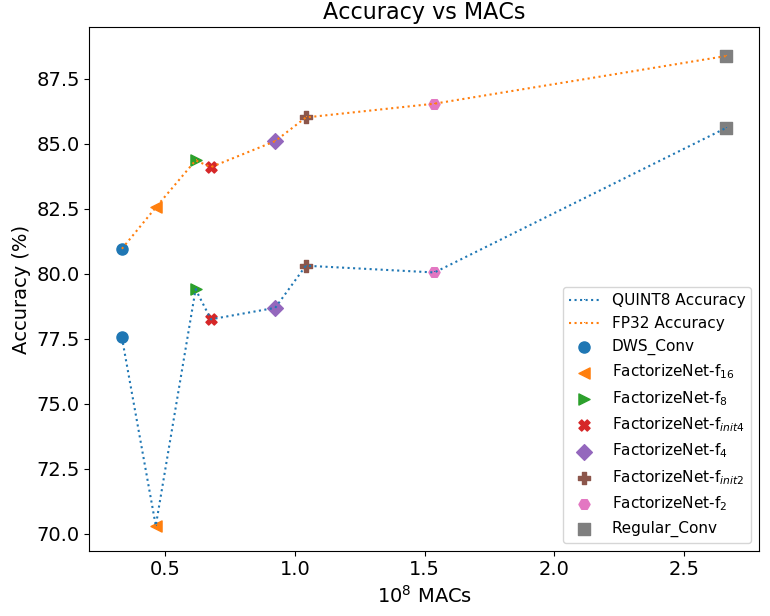} 
    \end{minipage}\hfill
    \centering
    \begin{minipage}{0.25\textwidth}
        \centering
        \includegraphics[width=0.99\textwidth]{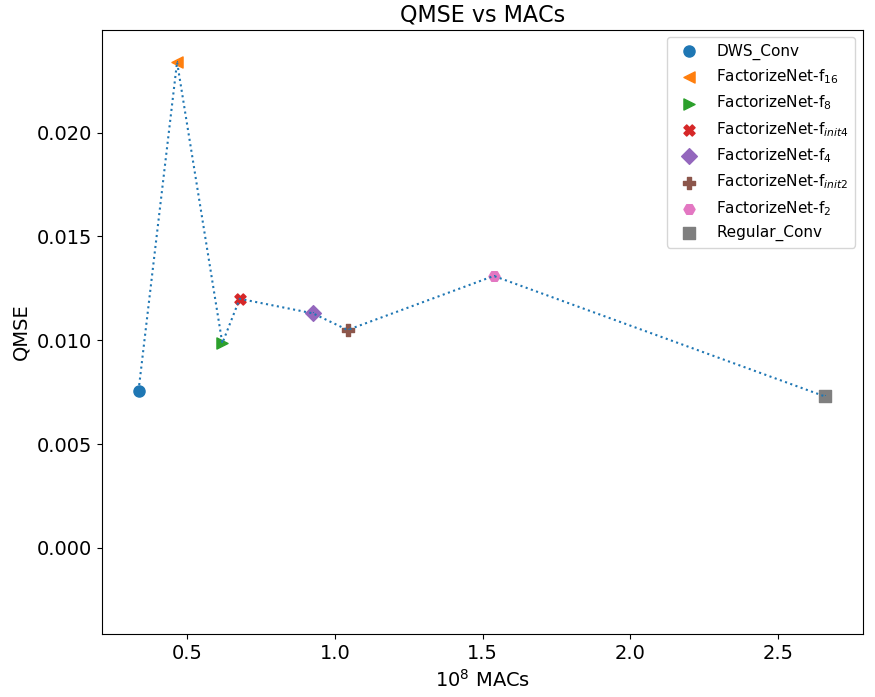} 
    \end{minipage}\hfill
    \begin{minipage}{0.25\textwidth}
        \centering
        \includegraphics[width=0.99\textwidth]{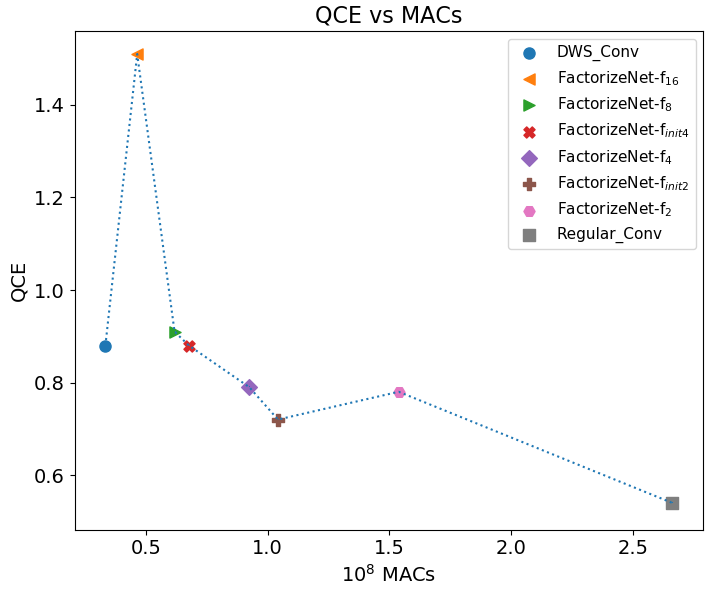} 
    \end{minipage}\hfill
    \begin{minipage}{0.25\textwidth}
        \centering
        \includegraphics[width=0.95\textwidth]{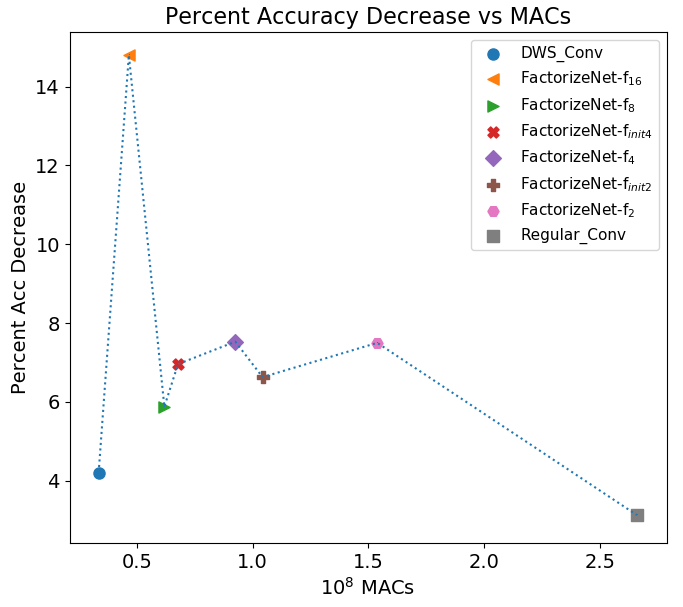} 
    \end{minipage}
    \caption{\footnotesize{} Best viewed in colour. \textbf{Far left}: Accuracy vs MACs (fp32 and quint8 accuracy) under depth factorization. Since the Dense layers are fixed, we only compare the MAC totals of the convolution layers. \textbf{Center left}: QMSE vs MACs. \textbf{Center right}: QCE vs MACs. \textbf{Far right}: Percent accuracy decrease vs MACs.}
\label{fig:quantizationerrormetrics}
\vspace{-0.15in}
\end{figure*}

\vspace{-0.15in}
\section{Experiment}
\vspace{-0.15in}
We start with a VGG-like macroarchitecture (see Figure~\ref{fig:factorizenet}) trained and tested on CIFAR-10. As we begin to factorize, the regular convolution layers (except for the first layer, which stays constant) are replaced with ``Groupwise Separable" Convolution where factorization rate \textit{f} is a programmable parameter. We refer to the resulting set of architectures as FactorizeNet. The groupwise separable convolution follows the structure of depthwise separable convolutions~\cite{DBLP:journals/corr/HowardZCKWWAA17}. Ie. GroupConv-BatchNorm-Relu-PointwiseConv. When \textit{f = input depth}, we recover depthwise separable convolutions. Following best practices, we always use a Conv-BatchNorm-Relu op-pattern. We demonstrate two progressively increasing factorization methods. The first is a uniform factorization configuration. Ie. A single factorization rate is applied to every Groupwise Separable Conv layer in the network. We progressively double this factorization rate on each step through the search space. We train networks with uniform factorizations of \textit{f = 2, 4, 8, 16}. These networks are denoted FactorizeNet-f\textsubscript{j} where \textit{j} is uniform factorization rate (e.g., FactorizeNet-f\textsubscript{2} is the network with a uniform factorization rate of 2). The second approach is to progressively double the factorization rate as we go deeper into the CNN in a Reverse Pyramid configuration (see Figure~\ref{fig:reversepyramid} for details). For Reverse Pyramid factorization, we train networks with \textit{f\textsubscript{init} = 2, 4}. These networks are denoted FactorizeNet-f\textsubscript{initk} where \textit{k} is initial factorization rate (eg. FactorizeNet-f\textsubscript{init2} is the network with reverse pyramid factorization and initial factorization rate of 2). We also train FactorizeNet with regular convolution and depthwise separable convolution in place of Groupwise Separable Conv (denoted Regular\_Conv and DWS\_Conv). Each network is trained from scratch for 200 epochs of SGD with Momentum = 0.9, batch-size = 128, and Glorot Uniform initializer~\cite{GlorotInit} for all layers. Initial learning rate is 0.01 and we scale it by 0.1 at the 75th, 120th, and 170th epochs. For the activation range tracking we perform top/bottom 1\% clipping computed on a random sample of 1024 training samples. Basic data augmentation includes vertical/horizontal shift, zoom, vertical/horizontal flip and rotation. We use Tensorflow for training and quantizing the weights and activations to quint8 format. Basic top/bottom 1\% percentile clipping is used for activation quantization as it is a common, low-overhead method.

For each network we observe the efficiency-accuracy trade-offs with respect to 4 quantities: fp32 accuracy, quantized 8-bit (quint8) accuracy, quantized mean-squared error (QMSE), and quantized crossentropy (QCE). QMSE refers to the MSE between the fp32 network outputs and the quint8 network outputs after dequantization. Similarly, QCE measures the cross entropy between the fp32 network outputs and the dequantized quint8 network outputs. While QMSE directly measures the difference in network output, QCE quantifies the difference in distribution of the network outputs. For classification, QCE can sometimes be more reflective of differences in behaviour. Additionally, we also observe the relative accuracy degradation (change in accuracy divided by fp32 accuracy) of each network after quantization. Figure~\ref{fig:quantizationerrormetrics} shows these quantities vs MAC-count.

\begin{figure*}
\vspace{-0.12in}
    \centering
    \begin{minipage}{0.31\textwidth}
        \centering
        \includegraphics[width=0.9\textwidth]{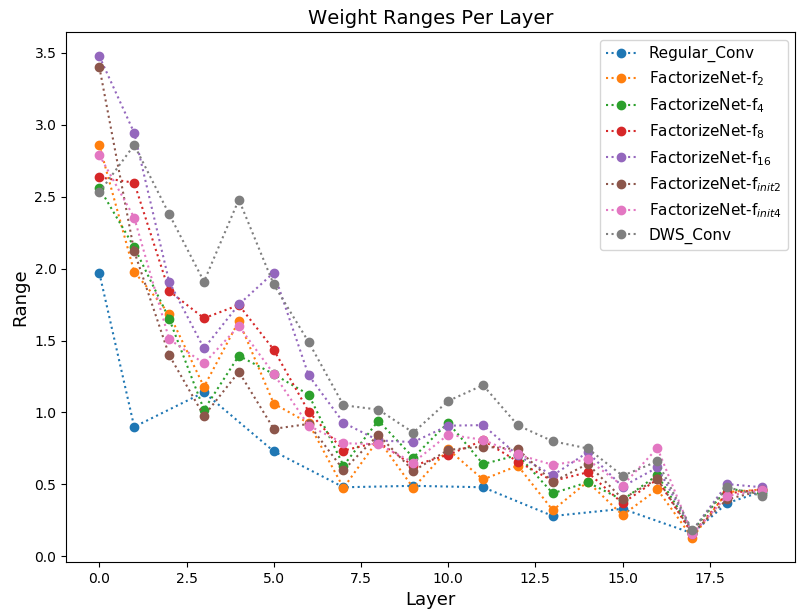} 
    \end{minipage}\hfill
    \begin{minipage}{0.31\textwidth}
        \centering
        \includegraphics[width=0.9\textwidth]{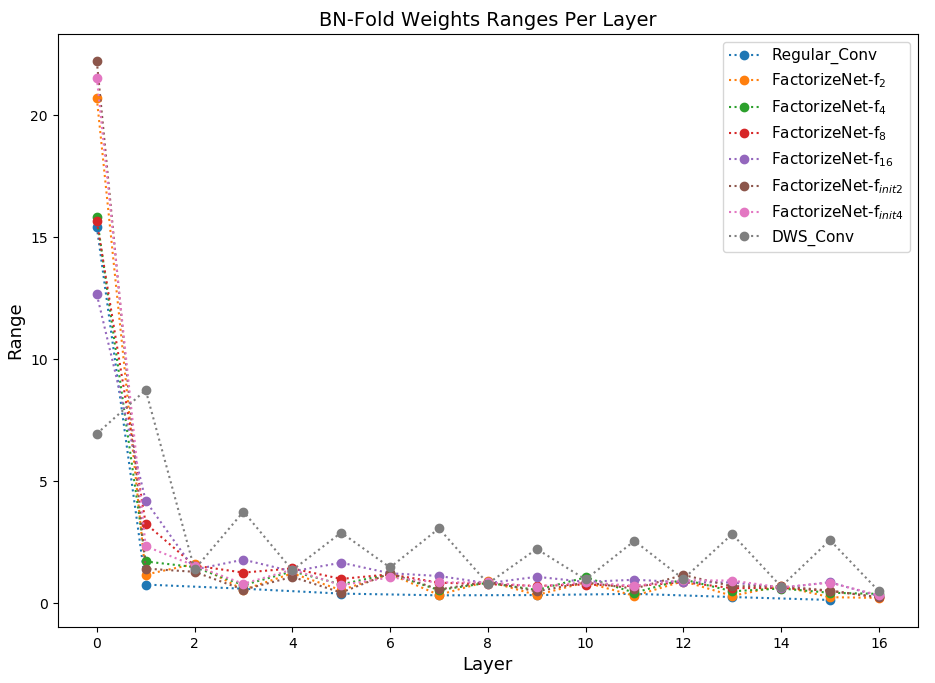}
    \end{minipage}\hfill
    \begin{minipage}{0.32\textwidth}
        \centering
        \includegraphics[width=0.9\textwidth]{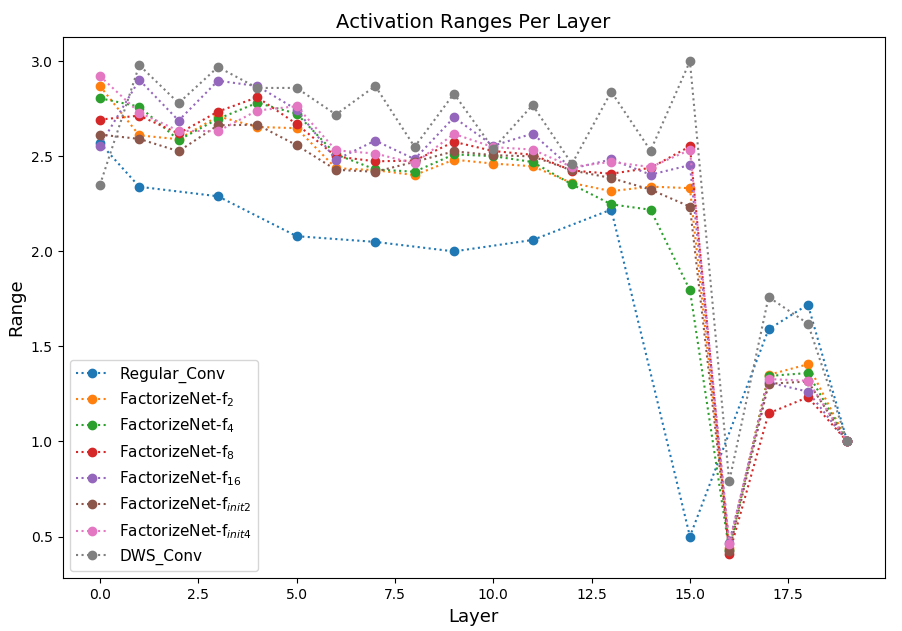} 
    \end{minipage}
    \caption{\footnotesize{} \textbf{Left}: Weights ranges per layer. \textbf{Center}: BN-Fold weights ranges. \textbf{Right}: Activations ranges. \textbf{Note:} Due to lack of space, we did not show the average precisions. However, these are still valuable statistics.}
    \vspace{-0.05in}
\label{fig:layerwise-data}
\end{figure*}

\begin{figure*}
\vspace{-0.05in}
    \centering
    \begin{minipage}{0.23\textwidth}
        \centering
        \includegraphics[width=0.96\textwidth]{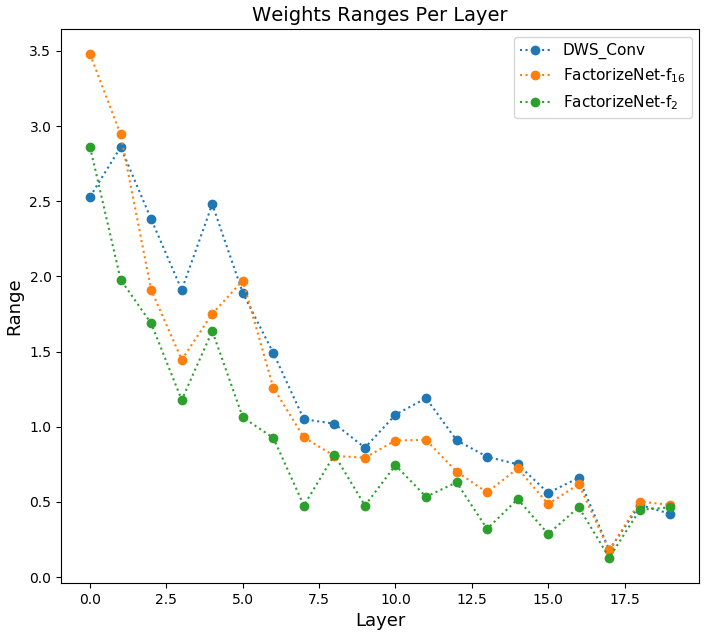} 
    \end{minipage}\hfill
    \begin{minipage}{0.23\textwidth}
        \centering
        \includegraphics[width=0.95\textwidth]{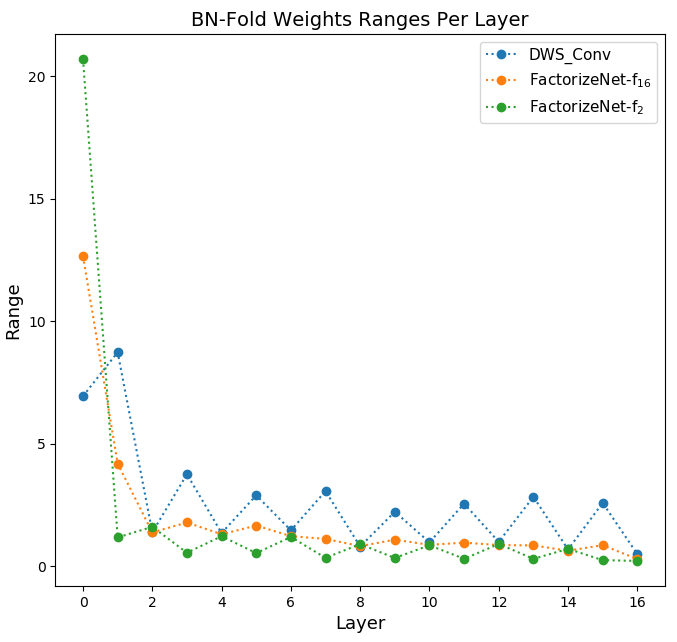} 
    \end{minipage}\hfill
    \begin{minipage}{0.23\textwidth}
        \centering
        \includegraphics[width=0.98\textwidth]{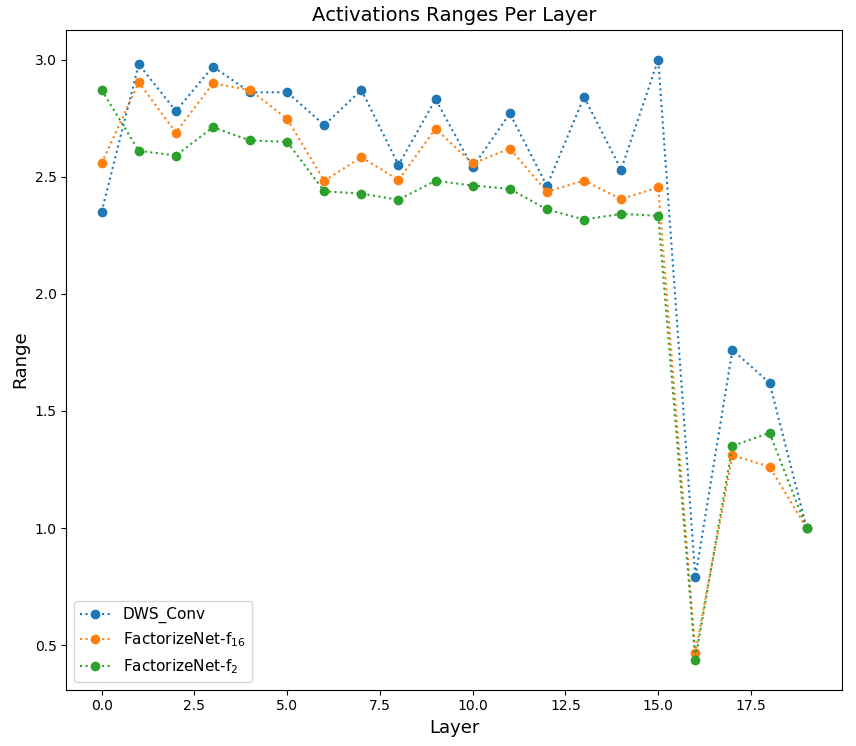} 
    \end{minipage}\hfill
    \begin{minipage}{0.23\textwidth}
        \centering
        \includegraphics[width=0.98\textwidth]{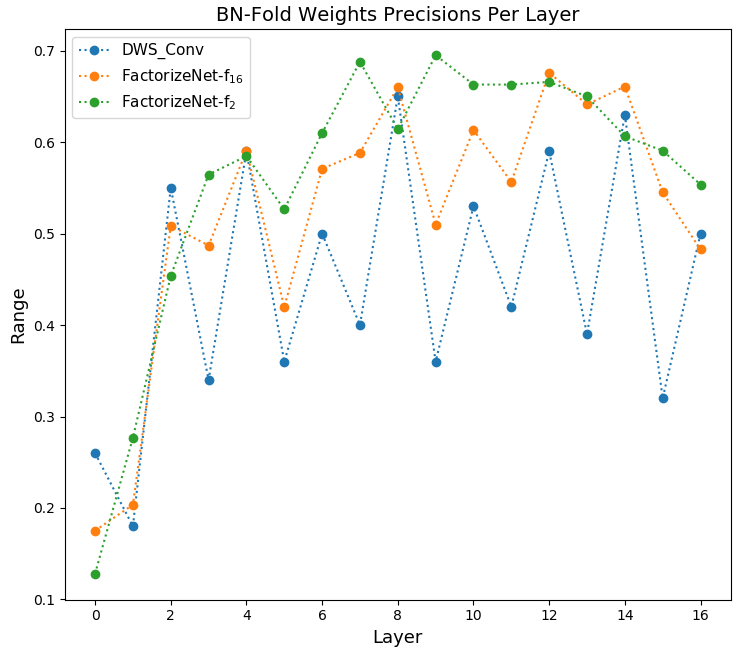} 
    \end{minipage}
    \caption{\footnotesize{}Comparing DWS\_Conv, FactorizeNet-f\textsubscript{16}, and FactorizeNet-f\textsubscript{f2} \textbf{Far left}: Weights ranges. \textbf{Center left}: BN-Fold weights ranges. \textbf{Center right}: Activations ranges. \textbf{Far right}: BN-Fold weights precisions.}
    \vspace{-0.22in}
\label{fig:focus-analysis}
\end{figure*}

\vspace{-0.12in}
\section{Discussion}
\vspace{-0.12in}
\label{sec:results}

From Figure \ref{fig:quantizationerrormetrics}, we have a high-level picture of the efficiency/accuracy trade-offs. Interestingly, FactorizeNet-f\textsubscript{init2} (104.3 MMACs, 86.01\% fp32 acc, 80.31\% quint8 acc) has less MACs than FactorizeNet-f\textsubscript{2} (153.8 MMACs, 86.54\% fp32 acc, 80.05\% quint8 acc) but similar accuracy. Furthermore, if targeting fp32 environments, FactorizeNet-f\textsubscript{init2} would offer over 2.5x MAC reduction from Regular\_Conv (266.0 MMACs, 88.37\% fp32 acc, 85.60\% quint8 acc) with a very small accuracy reduction. When analyzing quantized accuracy, some interesting anomalies emerge. Specifically the sharp drop in accuracy for FactorizeNet-f\textsubscript{16} (14.8\% relative accuracy drop). Also worth noting is that while most of the other models have higher quantized accuracy, DWS\_Conv experiences a noticeably smaller \textit{relative decrease} in quantized accuracy (4.21\% vs. 5.88\% - 7.53\%). This may be due to the much smaller increase in range of the BN-Fold weights in its first layer.

To get a better understanding of the factors contributing to the degradation in
FactorizeNet-f\textsubscript{16}, we move to our low-level analysis. Figure~\ref{fig:layerwise-data} shows the dynamic ranges of each layer. This low-level information gives us a direct look at the underlying distributions and how they interact with quantization noise. For example, besides generally smaller weights ranges (both convolution weights and batchnorm-folded weights), Regular\_Conv activations ranges are also noticeably lower. This begins to explain why Regular\_Conv is so robust to quantization (3.13\% relative accuracy loss). Going back to FactorizeNet-f\textsubscript{16}, the increased BN-Fold weights ranges early in the network may begin to explain why this CNN experienced a sharp drop in quantized accuracy. Furthermore, if we analyze the average precision of the BN-Fold weights in FactorizeNet-f\textsubscript{16} we see a combination of large range and low precision in the early, low-level feature extraction layers. Interestingly, the BN-Fold weights in FactorizeNet-f\textsubscript{2} show an even worse average precision in the first layer. However, the precision of BN-Fold weights in FactorizeNet-f\textsubscript{2} is higher on average and hints at a more representative projection of the network’s layers from their continuous distribution into a discretized space. Furthermore, we observe a generally lower range of activations for Factorizenet-f\textsubscript{2}. See Figure~\ref{fig:focus-analysis} for detailed comparison. Zooming back out to the inter-network trends, we can see from the BN-Fold weights ranges that there may be a significant loss of information in the early low-level feature extraction stages. It would be interesting to see how these distributions change if we do not use BatchNorm for the first layer since the pre-BN-Fold weights have a much smaller range. While it is intractable to pinpoint any single reason for the observed quantized behaviour, our layer-level analysis reveals a rich set of interconnected factors contributing to each network’s system dynamics. We could even further expand our analysis to use more rigorous, yet scalable statistical methods for layerwise analysis. From these initial analyses, we see that a fine-grained, systematic analysis can yield detailed insights to help further guide our design process.

\vspace{-0.17in}
\section{Conclusion}
\vspace{-0.15in}
\label{sec:conclusion}

We introduce a systematic, progressive depth factorization strategy coupled with a fine-grained layerwise analysis for exploring the efficiency/accuracy trade-offs of factorizing CNN architectures. In doing so, we can gain detailed insights on the impact of depth factorization on final floating point and quantized accuracy and also identify the optimal factorization configuration (ie. FactorizeNet). Future work includes using more sophisticated algorithms for increasing factorization, investigating activation sparsity under factorization, and factorizing more complex blocks/architectures.

\vspace{-0.17in}

\bibliographystyle{IEEEtran}
\footnotesize{}
\vspace{-0.02in}
\bibliography{emc2_depth_factorization.bib}
\end{document}